\documentclass[conference]{IEEEtran}
\IEEEoverridecommandlockouts
\usepackage{amsmath,amssymb,amsfonts}

\usepackage{tcolorbox}
\usepackage{longtable}
\usepackage{booktabs}
\usepackage{hyperref}

\usepackage{graphicx}

\begin{document}

\title{Mapping Trustworthiness in Large Language Models: A Bibliometric Analysis Bridging Theory to Practice}

\author{

\IEEEauthorblockN{José Antonio Siqueira de Cerqueira}
\IEEEauthorblockA{
Tampere University\\
Tampere, Finland \\
jose.siqueiradecerqueira@tuni.fi} \\

\IEEEauthorblockN{Nannan Xi}
\IEEEauthorblockA{
Tampere University\\
Tampere, Finland \\
nannan.xi@tuni.fi}

\and

\IEEEauthorblockN{Kai-Kristian Kemell}
\IEEEauthorblockA{
Tampere University\\
Tampere, Finland \\
kai-kristian.kemell@tuni.fi}\\

\IEEEauthorblockN{Juho Hamari}
\IEEEauthorblockA{
Tampere University\\
Tampere, Finland \\
juho.hamari@tuni.fi}

\and

\IEEEauthorblockN{Rebekah Rousi}
\IEEEauthorblockA{
University of Vaasa\\
Vaasa, Finland \\
rebekah.rousi@uwasa.fi}\\

\IEEEauthorblockN{Pekka Abrahamsson}
\IEEEauthorblockA{
Tampere University\\
Tampere, Finland \\
pekka.abrahamsson@tuni.fi}

}

%the shift on focus from ai ethics to llm trustworthiness has made researchers to repeat the same mistake of spreading new guideliness, this time rebranded, as it was seen it was a mistake before because of no real outcome, this is stil leading to ethics washing, where indivual proposals aims to ease public criticisms and resist to legislations while no real commitment is done.

\maketitle

\thispagestyle{plain}

\begin{abstract}
The rapid proliferation of Large Language Models (LLMs) has raised significant trustworthiness and ethical concerns. Despite the widespread adoption of LLMs across domains, there is still no clear consensus on how to define and operationalise trustworthiness. This study aims to bridge the gap between theoretical discussion and practical implementation by analysing research trends, definitions of trustworthiness, and practical techniques. We conducted a bibliometric mapping analysis of 2,006 publications from Web of Science (2019-2025) using the Bibliometrix, and  manually reviewed 68 papers. We found a shift from traditional AI ethics discussion to LLM trustworthiness frameworks. We identified 18 different definitions of trust/trustworthiness, with transparency, explainability and reliability emerging as the most common dimensions. We identified 20 strategies to enhance LLM trustworthiness, with fine-tuning and retrieval-augmented generation (RAG) being the most prominent. Most of the strategies are developer-driven and applied during the post-training phase. Several authors propose fragmented terminologies rather than unified frameworks, leading to the risks of ``ethics washing,'' where ethical discourse is adopted without a genuine regulatory commitment. Our findings highlight: persistent gaps between theoretical taxonomies and practical implementation, the crucial role of the developer in operationalising trust, and call for standardised frameworks and stronger regulatory measures to enable trustworthy and ethical deployment of LLMs.
\end{abstract}


\section{Introduction}
Artificial Intelligence (AI) has become a cornerstone of technological progress, yet remains at the centre of ethical debate. More than 200 ethical guidelines and principles have been proposed to guide its development and use, reflecting the growing recognition of its societal impact \cite{hagendorff2020ethics, Correa2023WorldwideAIethics200}. These guidelines, proposed by public organisations, society, academia and private companies, do not provide objective guidance to developers, but rather abstract principles that are not practical for developers to operationalise, e.g, transparency, privacy, justice \& fairness \cite{Vakkuri2021ECCOLAjournal, floridi2022unified, Cerqueira2022Guide}.

As an evolution of AI system that employs advanced deep learning techniques, Large Language Models (LLMs) are trained on vast datasets and capable of generating novel text based on the data it was trained on \cite{sun2024trustllm}. Their rapid adoption has sparked both enthusiasm and concern, as their potential for negative outcomes -- e.g., spreading misinformation, perpetuating biases, hallucination -- has become evident \cite{liu2023trustworthy, Wang2023DecodingTrustAC}. 

%Despite efforts like the Paris AI Action Summit \cite{elysee_2025}, which seeks to promote responsible AI development and international cooperation, significant challenges remain. For instance, Google’s recent decision to remove its AI principles prohibiting the use of AI in autonomous weapons and surveillance has raised questions about the commitment of major tech companies to ethical AI \cite{google2025endingaiban}. This shift underscores the tension between ethical ideals and commercial or geopolitical interests, highlighting the need for robust, enforceable legislations.

The swift popularization of LLMs have encouraged researchers in understanding the specific ethical dimensions of this new generative AI (genAI), introducing them mainly as trustworthiness taxonomies, i.e., a rebranded set of similar AI ethical principles \cite{sun2024trustllm, liu2023trustworthy, Wang2023DecodingTrustAC}. Trustworthiness in LLMs can be characterized in terms of reliability, safety, fairness, resistance to misuse, explainability \& reasoning, social norm and robustness \cite{liu2023trustworthy}. Thus, the focus of AI ethics discussions has moved to the trustworthiness of LLM systems. The current debate on LLM trustworthiness largely focus on providing definitions, benchmarks and evaluation methods \cite{sun2024trustllm, liu2023trustworthy, Wang2023DecodingTrustAC}.

Analogously with AI ethics, the proliferation of trustworthiness taxonomies for LLM has not led to a clear consensus on definitions or a practical framework to guide the development of these systems \cite{smith2025responsiblegenerativeaiuse}. Despite the growing body of literature on LLM trustworthiness, there is still no unified set of practical guidelines to help developers of LLM-based systems mitigate risks throughout the development and deployment process. This gap between theory and practice not only hinders the creation of more trustworthy LLM systems but also masks deeper structural problems that have persisted since the early days of AI ethics. While only a limited number of actors develop foundational LLMs, the widespread integration of these models into downstream systems highlights the urgent need for trustworthiness-enhancing practices that can be applied by practitioners leveraging on LLMs. This study aims to bridge the gap between theory and practice by investigating key research trends, definitions of LLM trustworthiness, and practical means of operationalising it found in the literature.

We conducted a bibliometric mapping of trust, trustworthiness, and ethics in LLMs using the Web of Science database. An initial set of 2,006 studies published between 2019 and 2025 was collected and analyzed to identify influential authors, key research trends, and emerging approaches.
To deepen our analysis, we filtered a subset of 68 papers that explicitly mentioned ``trust*'' in their abstracts. These papers were manually reviewed to identify definitions and practical strategies that developers can use to enhance trustworthiness across the LLM development lifecycle. By bringing these perspectives together, we aim to provide both a theoretical and practical basis for promoting more ethical and trustworthy LLMs.

In order to guide this study, we devised the following Research Questions:
\begin{itemize}
    \item RQ1: What are the research trends, key authors, and thematic areas in the study of LLM trustworthiness?
    \item RQ2: What are the LLM trust/trustworthiness definitions proposed by different authors in the literature?
    \item RQ3: What practical techniques and tools can be used to enhance LLM trustworthiness throughout the LLM lifecycle?
\end{itemize}

%Tentative
The main contributions of this study are as follows:
\begin{itemize}
\item AI ethics research focus has shifted from abstract guiding principles to abstract LLM trustworthiness taxonomies. \textbf{Several authors propose fragmented terminologies rather than unified frameworks, leading to the risks of ``ethics washing,'' where ethical discourse is adopted without a genuine regulatory commitment.}

\item We compiled 18 definitions of trust / trustworthiness in LLMs proposed by different authors. Transparency, explainability and reliability are the most frequent terms associated. \textbf{Most of reviewed papers build on Mayer et al.'s \cite{Mayer1995} definition of organizational described as a three factor construct: ability, capacity to perform tasks effectively; benevolence, intent to act in users' best interest; integrity, adherence to ethical stands.}

\item We identified 20 strategies for enhancing trustworthiness in LLMs. \textbf{Most strategies occur during the post-training phase and are developer-driven, with supervised fine-tuning, retrieval-augmented generation (RAG), and knowledge graphs being the most cited techniques. Developers are responsible for the majority of these strategies (16 out of 20), highlighting an imbalance in accountability and underscoring their central role in operationalising trustworthiness.}

\end{itemize}

%Through the combination of a bibliometric analysis 

%By integrating a systematic review of trustworthiness definitions with bibliometric insights and practical techniques, this work contributes to ongoing efforts to operationalize trustworthiness in LLMs and offers concrete steps toward the development of more trustworthy LLM systems.

%This article is organized as follows: in the Section \ref{related_work_section}, we present studies that explore LLM trustworthiness taxonomies, and conduct similar reviews and present some AI ethical guidelines studies, highlighting their similarities and differences. In Section \ref{methodology_section}, we describe the bibliometric analysis methodology adopted through the use of Bibliometrix tool. In Section \ref{results_section}, we present the main findings, including systematic analyses of LLM trustworthiness definitions and trustworthiness-enhancing techniques. In Section \ref{limitations}, we discuss limitations, in Section \ref{discussion_and_conclusions}, we summarise main findings and implications, suggesting future directions for research and development.

\section{Related Work}
\label{related_work_section}

%Here just discuss other literature reviews related to AI ethics. You can just build on "related work" from this and modify slightly: https://helda.helsinki.fi/server/api/core/bitstreams/6996a645-6972-472d-87f7-7b60349e37b9/content

Several authors have analysed AI ethical guidelines, proposing novel sets of principles, such as \cite{jobin2019globallandscape, Correa2023WorldwideAIethics200, hagendorff2020ethics, Ryan2020ArtificialIE}. Correa et al. \cite{Correa2023WorldwideAIethics200} performed a meta-analysis of 200 AI ethics documents, summarising them in 17 principles, where 5 principles emerge as more recurrent: 1) Transparency / Explainability / Auditability, 2) Reliability / Safety / Security / Trustworthiness, 3) Justice / Equity / Fairness/ Non-discrimination, 4) Privacy and 5) Accountability / Liability. Morley et al. \cite{Morley2019} provide an overview of 106 publicly available tools designed to help operationalize ethical principles in AI. Based on five principles -- beneficence, non-maleficence, autonomy, justice, and explicability -- the authors map practical techniques to the machine learning lifecycle phase , illustrating how these tools can support ethical AI development in practice. Studies on AI ethics that emerged before the rise of LLMs and generative AI offer a useful starting point for exploring trustworthiness in these technologies. They provide the background and foundational frameworks on which future research into LLM trustworthiness can build, since LLMs are ultimately part of the broader field of AI.

Regarding LLM and genAI trust/trustworthiness, recent studies are proposing similar set of theoretical taxonomies along with evaluation methods against such taxonomies. Liu et al. \cite{liu2023trustworthy} propose a taxonomy for evaluating trustworthiness in LLM in seven major categeories: reliability, safety, fairness, resistance to misuse, explainability and reasoning, adherence to social norms, and robustness. Although the paper is not a literature review, it does describe each of the major categories with examples. However, the authors fail to provide a clear definition of what is trustworthy and do not provide practical guidance for developers on how to operationalise their taxonomy. Sun et al. \cite{sun2024trustllm} and Wang et al. \cite{Wang2023DecodingTrustAC} propose different sets of principles for trustworthiness in LLM, such as truthfulness, safety, fairness, robustness, privacy, machine ethics, transparency, accountability. Nonetheless, they do not provide developers with a practical guidance on how to operationalise these principles.

Dhoska and Bebi \cite{Dhoska2025} performed a bibliometric analysis of \textit{trustworthy in AI} in the Scopus database, showing e.g., relevant authors, affiliations, countries and most cited papers. Nevertheless, their study focused on AI rather than LLM specifically, and they did not present the importance of trustworthiness in AI in their discussion, focusing only on the results of the bibliometric metrics and not on the content of the papers analysed. Albahri et al. \cite{Albahri2023} conducted a systematic science mapping analysis in the context of trustworthy and explainable AI (XAI) in healthcare systems, and found various XAI techniques for developers to operationalise. Unlike their approach, we aim to provide means to operationalise more of the trustworthiness categories.

Despite the aforementioned efforts in this area, our work differs by exploring not only a bibliometric analysis and definitions of trustworthiness in LLM, but also practical means for developers to improve LLM trustworthiness throughout the LLM lifecycle.

\section{Methodology}
\label{methodology_section}

Bibliometric analysis is a quantitative research method used to identify trends, influential works, and key research themes in a given domain \cite{Donthu2021HowTCbibliometric}. To achieve this, we utilize Bibliometrix, an open-source R package specifically designed for bibliometric and scientometric studies \cite{aria2017bibliometrix}. We extracted data from \href{https://www.webofknowledge.com)}{Web of Science (WoS) Core Collection} during 21st of February 2025, as this database contain high-quality, peer-reviewed publications in AI and Software Engineering, and provides high quality of publications and reliability in its indexing of high-ranking journals \cite{Caputo2021DigitalizationAB}. The search query was formulated using a combination of relevant broad keywords seeking for a comprehensive dataset:

\begin{tcolorbox}

TS=((``Large Language Model*'' OR LLM OR ``Generative AI'' OR ``genAI'') AND (Trust* OR Ethic*))

\end{tcolorbox}

Where TS stands for Topic, which encompasses the following fields within a record: Title, Abstract, Author Keywords and Keywords Plus®. The latter is automatically generated by the database, in which its data are words or phrases that frequently appear in the titles of an article's references, but do not appear in the title of the article itself \footnote{\href{https://support.clarivate.com/ScientificandAcademicResearch/s/article/KeyWords-Plus-generation-creation-and-changes?language=en_US}{https://support.clarivate.com/ScientificandAcademicResearch/s/article/KeyWords-Plus-generation-creation-and-changes}}.

The inclusion criteria for the retrieved publications were:
\begin{itemize}
    \item Peer-reviewed journal articles and conference proceedings.
    \item Papers published in the last \textbf{6 years} (2019-2025).
    \item Studies encompassing discussion on trustworthiness or ethics of LLMs or generative AI.
    \item Articles written in \textbf{English}.
\end{itemize}

\textbf{Bibliometric analysis:} After filtering out duplicates, irrelevant papers and LLM as an author abbreviation, we obtained a \textbf{final dataset of 2,006 publications}. This number of studies discussing LLMs is remarkable as we reduced the year range of our search to only 6 years due to the fact that LLM is a new technology. A BibTex file was extracted from WoS and imported into Bibliometrix, where we carried out the \textbf{bibliometric analysis to answer RQ 1.}

\textbf{Further manual reading:} After exploring the broader set with Bibliometrix, we conducted a manual reading of a reduced dataset. In order to obtain more insightful information related to our research focus, we filtered papers that had ``trust*'' in the abstract. Thus, \textbf{a subset of 68 papers was obtained, and we used this subset to answer RQs 2 and 3.}

\section{Results}
\label{results_section}

This study examines the landscape of trust, trustworthiness and ethics research in LLM and genAI from 2019 to 2025. A total of 2,006 relevant studies published in 1,160 different venues over the past six years were analysed. These publications were authored by 7,901 researchers, with an average citation count of 8.901 per document. Notably, collaborative research dominates the field, with 96.6\% of the studies involving multiple authors (7,632 in total), while only 3.4\% were produced by single authors (269 individuals).

% Here was RQ1

\begin{figure}[htbp]
    \centering
    \includegraphics[width=\linewidth]{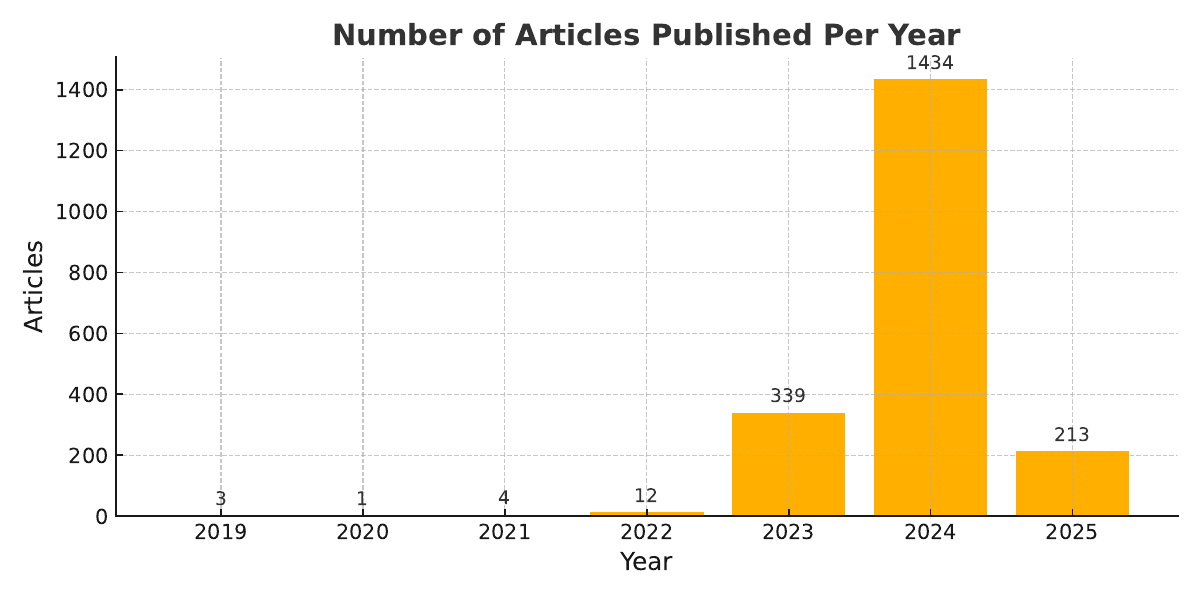}
    \caption{Evolution of Studies Per Year - 2019 - 2025.}
    \label{fig:annualscientific_production}
\end{figure}

%%%%%%%%%%%%%%%%%%%%%%%%%%%%%%%%%%
% Anual Publication Per Year
The study shows a sharp rise in publications on LLM trustworthiness and ethics from 2019 to early 2025, growing from 3 articles in 2019 to 1,434 in 2024, as shown in Figure \ref{fig:annualscientific_production}. By February 2025, 213 articles had already been published, indicating continued momentum (~1,278 projected by year-end). Collaborative research dominated, with 96.6\% of studies involving multiple authors.

%%%%%%%%%%%%%%%%%%%%%%%%%%%%%
% Most Global Cited Documents

\begin{figure}[htbp]
    \centering
    \includegraphics[width=\columnwidth]{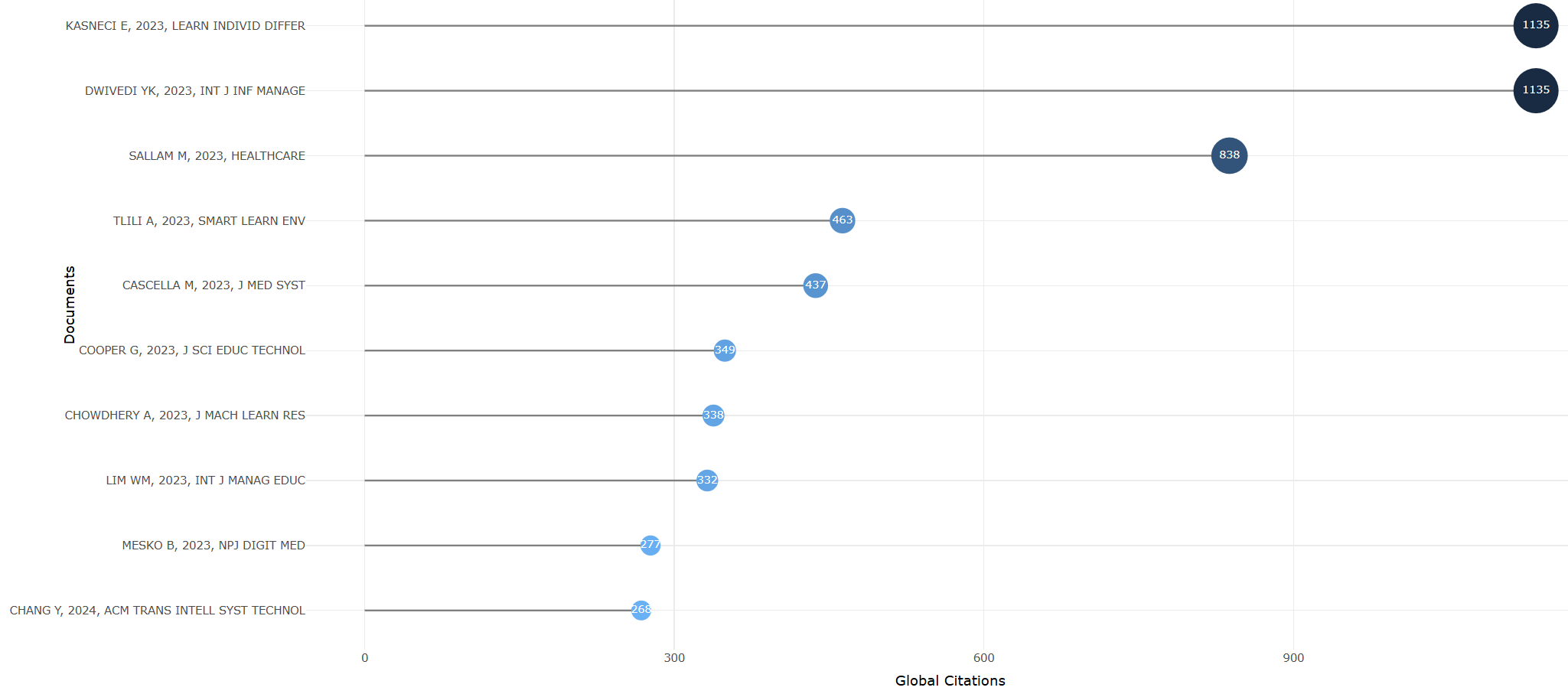}
    \caption{Most Globally Cited Articles (2022--2024).}
    \label{fig:most_Global_cited_documents}
\end{figure}

Figure \ref{fig:most_Global_cited_documents} shows the top articles ranked by total number of citations, without any filtering. The two most globally cited articles are ``ChatGPT for good? On opportunities and challenges of large language models for education'' by Kasneci et al. \cite{Kasneci2023} and ``Opinion Paper: “So what if ChatGPT wrote it?'' Multidisciplinary perspectives on opportunities, challenges and implications of generative conversational AI for research, practice and policy'' by Dwivedi et al. \cite{Dwivedi2023}, both with 1,135 global citations. The third most cited article ``ChatGPT Utility in Healthcare Education, Research, and Practice: Systematic Review on the Promising Perspectives and Valid Concerns'' by Sallam Malik \cite{Sallam2023}, with 838 global citations.

\begin{figure*}[t!]
    \centering
    \includegraphics[width=0.8\textwidth]{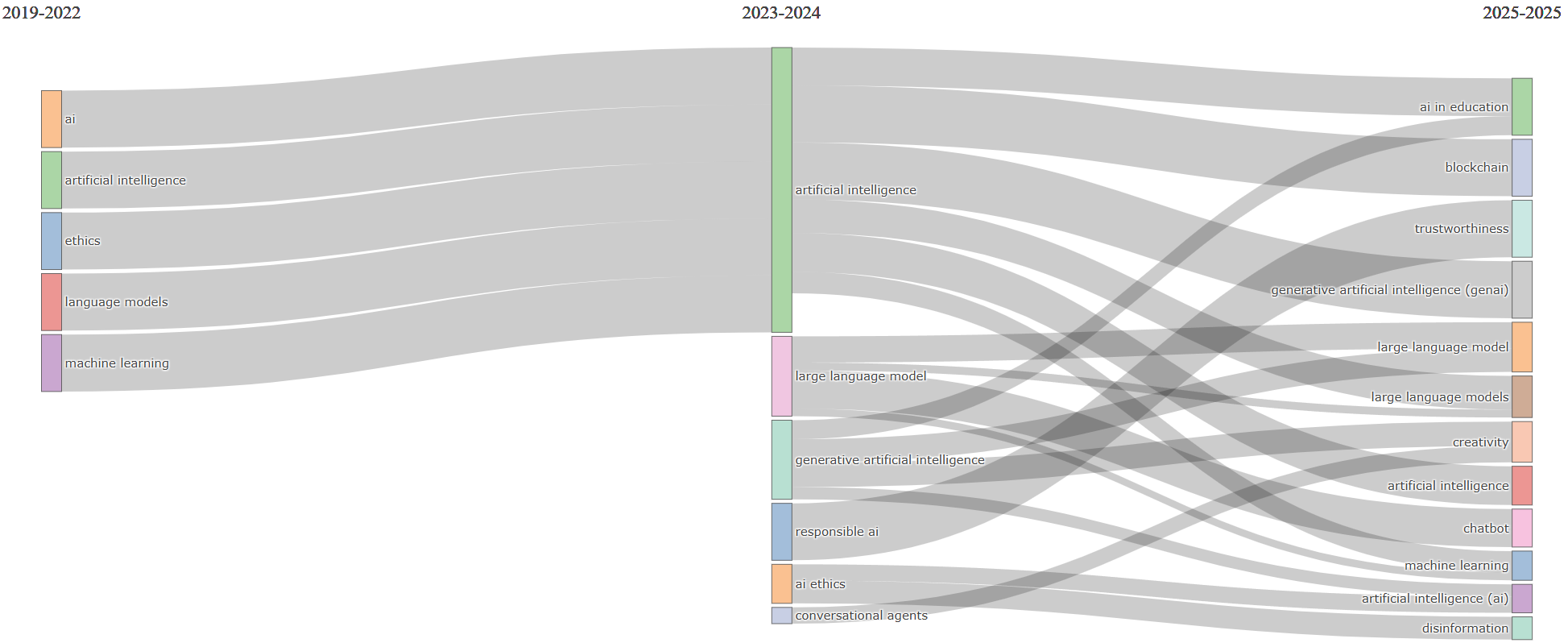}
    \caption{Thematic Evolution: 2019-2025.}
    \label{fig:thematic_evolution}
\end{figure*}

Thematic evolution of AI ethics, trustworthiness, and LLMs is shown in Figure \ref{fig:thematic_evolution}, which maps the shifts in research focus from 2019 to 2025. From 2019 to 2022, research primarily addressed broad topics such as artificial intelligence, ethics, machine learning, and language models, reflecting general concerns over AI development and ethical challenges.

However, in 2023-2024, the focus begins to narrow towards more specialised topics such as large language model, generative artificial intelligence, responsible AI and AI ethics. The emergence of these topics coincides with the surge in popularity of ChatGPT, but studies on the specific ethical or responsible dimensions of large language model and generative artificial intelligence are still in their infancy.

By 2025, the thematic scope diversified further, branching into AI in education, blockchain, trustworthiness, creativity, chatbots, and disinformation.  This evolution demonstrates the increasing complexity and multidisciplinary nature of the LLM trustworthiness debate.

%%%%%%%%%%%%%%%%%%%%%%%
% Word cloud

\begin{figure}[!h]
    \centering
    \includegraphics[width=\columnwidth, trim={0cm 30cm 0cm 30cm}, clip]{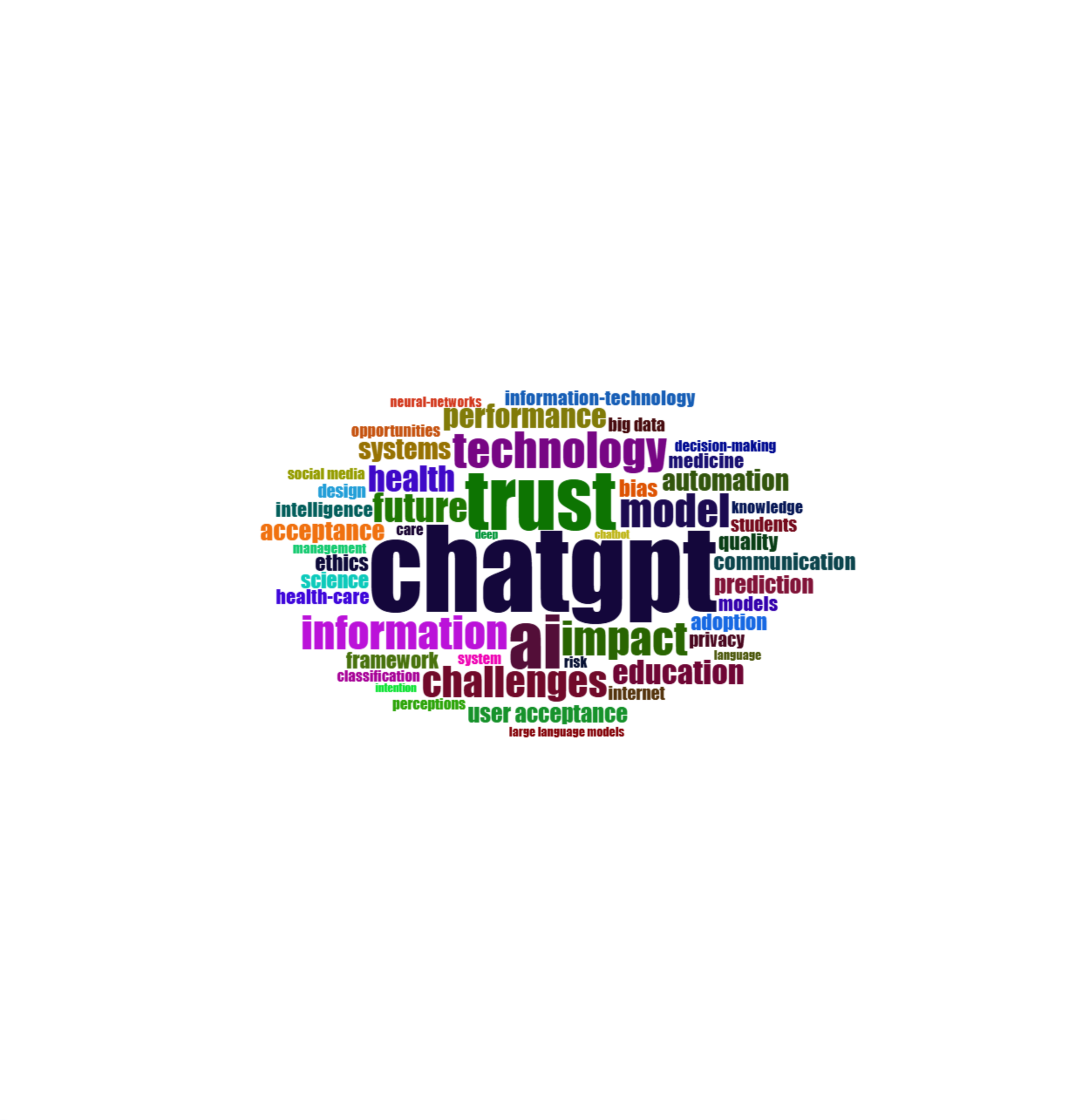}
    \caption{Most frequently occurring keywords in LLM trust and ethics literature (2019–2025)}
    \label{fig:word_cloud}
\end{figure}

The word cloud in Figure \ref{fig:word_cloud} represents the most frequently occurring keywords in the LLM trustworthiness and ethics literature. The size of each word corresponds to its relative frequency in the data set.

Notably, the terms ‘ChatGPT’, ‘trust’, ‘AI’, ‘technology’ and ‘information’ appear as the most dominant. Other common terms include ‘health’, ‘education’, ‘performance’, ‘impact’ and ‘user acceptance’. The presence of domain-specific terms such as ‘health' and ‘education' indicates relevant domains where LLMs are applied. In addition, several ethical related keywords -- such as ‘privacy’, ‘ethics’, ‘bias’ and ‘challenges’ -- indicates the researchers interest in this growing area.

%%%%%%%%%%%%%%%%%%%%%%%%%%
% Co-Occurrence Network

% The co-occurrence network visualization in Figure \ref{fig:co_occurrence_network} highlights the central role of "artificial intelligence" in the red cluster, with strong associations to key themes like "ChatGPT", "chatbot", "large language models", related to the technologies themselves, and to "medicine", "science", "education", "health", relating to domains of interest; and "ethics" and "bias" emphisizing the ethical concerns. This suggests that AI research is deeply interconnected with societal and ethical challenges, particularly in LLMs and their impact on various domains.

% On the other hand, the blue cluster focuses on the concept of "trust", linking it to "acceptance", "user acceptance", "perceptions", "adoption", "quality" and "privacy". This indicates a growing interest in AI trustworthiness, highlighting it as in terms of user acceptance. That is, instead of having previous set of AI ethical guidelines governing AI use, development and deployment in a more holistic way, this suggests a shift in a more simplistic view of trust in AI is simply gauging the acceptance of a LLM by the user.

% %The co-occurrence network in Figure \ref{fig:co_occurrence_network} highlights the divide between ethical AI development and its trust/trustworthiness implications.

% \begin{figure}[!h]
%     \centering
%     \includegraphics[width=\columnwidth]{Images/co-occurrence-network.png}
%     \caption{Co-Occurrence Network.}
%     \label{fig:co_occurrence_network}
% \end{figure}

%%%%%%%%%%%%%%%%%%%%%%%%%
% Coupling Map

\begin{figure}[htbp]
    \centering
    \includegraphics[width=\columnwidth]{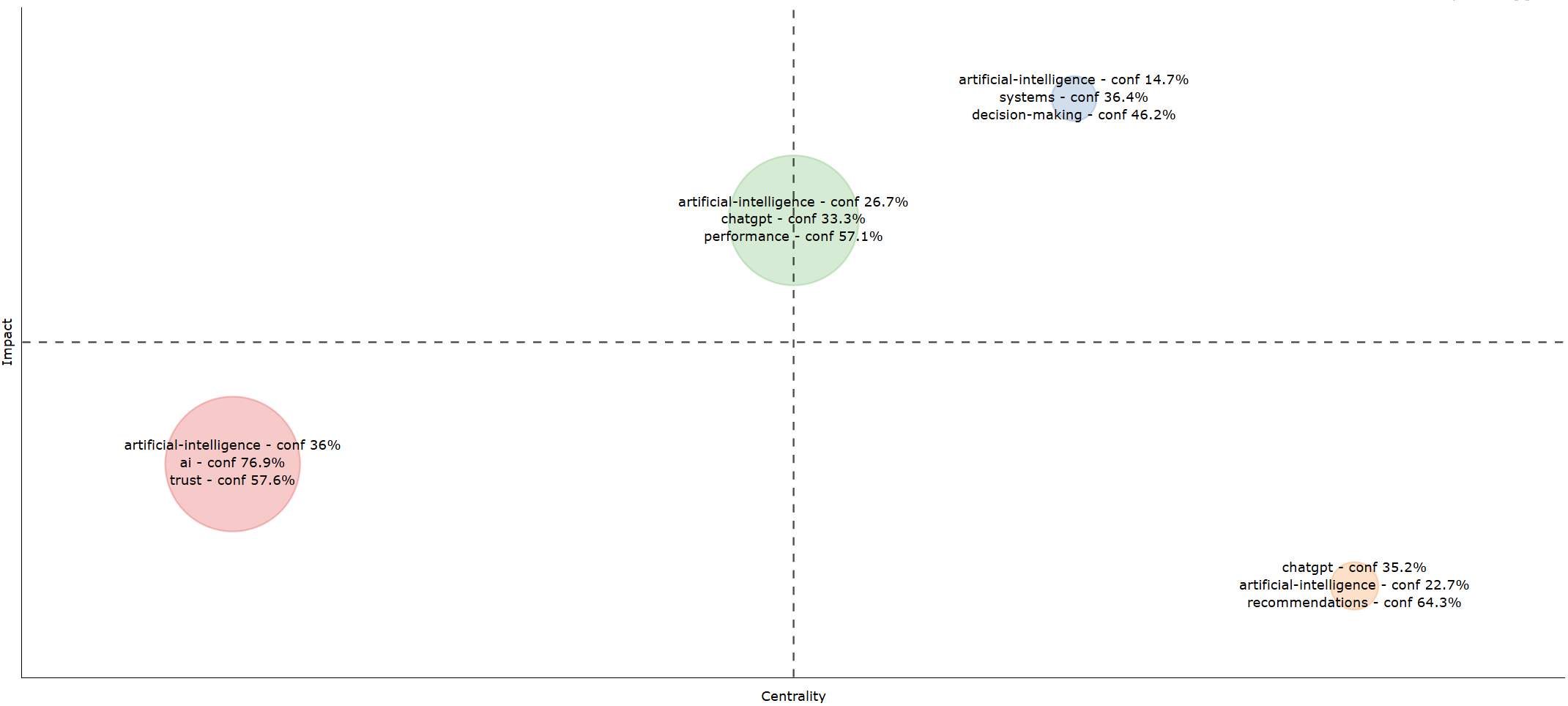}
    \caption{Clustering by Coupling.}
    \label{fig:clustering}
\end{figure}

We analyse studies that reference the same arrangement of cited articles in order to understand the bibliographic coupling. Through this analysis it is possible to cluster authors and identify possible themes. In Figure \ref{fig:clustering} four clusters are visible, highlighting the importance in academia of linking AI to 1) trust, 2) performance, 3) recommendations and 4) decision-making. Also noticeable is the appearance of ChatGPT in two clusters, in relation to performance and recommendations. Our main interest is focused on the red cluster, as it deals with trust. In this sense, we analyse studies from the five most prominent authors within this theme. Table \ref{tab:top_authors} shows the authors ranked by their relevance within the red cluster.

%The green cluster, positioned near the center, contains artificial intelligence, ChatGPT, and performance, indicating a balance between centrality and impact. This suggests that discussions about ChatGPT’s capabilities and AI performance are well-integrated within broader AI research. The top-right cluster, though smaller, highlights decision-making and AI systems, implying its emerging importance in AI governance. The analysis suggests a growing intersection between AI ethics, trust, and performance, reinforcing the need for responsible AI development and evaluation frameworks.

\begin{table}[h]
    \centering
    \renewcommand{\arraystretch}{1.2}
    \caption{Top Cited Authors within 'trust' Cluster and Their Works}
    \label{tab:top_authors}
    \resizebox{\columnwidth}{!}{%
    \begin{tabular}{clp{10cm}c}
        \toprule
        \textbf{\#} & \textbf{Author} & \textbf{Title} & \textbf{Citation} \\
        \midrule
        1  & Martinez-Maldonado R & Practical and ethical challenges of large language models in education: A systematic scoping review & \cite{Yan2023} \\
        2  & Park J & When ChatGPT Gives Incorrect Answers: The Impact of Inaccurate Information by Generative AI on Tourism Decision-Making & \cite{Kim2023} \\
        3  & Sallam M & ChatGPT Utility in Healthcare Education, Research, and Practice: Systematic Review on the Promising Perspectives and Valid Concerns & \cite{Sallam2023} \\
        4  & Wang C & A Survey on Evaluation of Large Language Models & \cite{chang2023survey} \\
        5  & Herrera F & Explainable Artificial Intelligence (XAI) 2.0: A manifesto of open challenges and interdisciplinary research directions & \cite{Longo2024} \\
        \bottomrule
    \end{tabular}
    }
\end{table}

%%%%%%%%%%%%%%%%%%%%%%%%%%%%%%%%%%%%%%%%%%%%%%%%%%%%%%%%%%%%%%%
% Table LLM Trustworthiness Definition
%\subsection{The LLM trust/trustworthiness definitions proposed by different authors in the literature}

% Here was RQ2

To address Research Questions 2 and 3, we manually analyzed 68 articles in which the term “trust*'' appeared in the abstract.

\begin{table*}[!t]
\centering
\caption{Trustworthiness Definitions in AI}
\label{tab:trust_definitions}
\renewcommand{\arraystretch}{1.1}
\resizebox{\textwidth}{!}{%
\begin{tabular}{|c|p{10cm}|p{3.2cm}|p{1.2cm}|p{3cm}|}
\hline
\textbf{\#} & \textbf{Trust/trustworthiness definition} & \textbf{Synonyms to trustworthy AI} & \textbf{Ref} & \textbf{Builds on} \\
\hline
1 & LLMs demonstrate strong generalization capabilities and adaptability to specific tasks through fine-tuning or in-context learning. However, they currently lack the predictability and robustness required for deployment 'as is' in adversarial environments. In this context, trustworthiness is characterized by the ability to be directly deployed in real-world scenarios while maintaining resilience against adversarial conditions. & Trustworthy, resilient, and interpretable artificial intelligence (AI) & \cite{Jha2023Challenges} & - \\
\hline
2 & To improve trust is to understand the behaviour of deep neural networks for Information Retrieval. & Explainable AI & \cite{Lucchese2023CanEmbeddings} & - \\
\hline
3 & Not explicitly, but states that: LLM’s ability in providing reliable security and privacy advice. & Trusted, Reliable & \cite{Chen2023CanLLMProvide} & - \\
\hline
4 & The perceived trustworthiness depends on several factors: the realism of the content, the context in which it is presented, and the pre-existing trust individuals have in the sources of information. Notably relevant to misinformation contexts. Addressing the ethical and political dimensions is crucial for maintaining trust in democratic societies. & Perceived trustworthiness, disinformation concerns & \cite{Carnevale2023-CARHEF-2} & - \\
\hline
5 & Trust in AI is described as the belief that an AI agent will support an individual in achieving their goals, particularly in situations marked by uncertainty and vulnerability. Trust plays a crucial role in user interaction with AI, as its absence can lead to reluctance in adopting AI tools, even when they demonstrate superior performance. Conversely, excessive trust in AI, especially in high-stakes domains like software engineering, can cause users to overlook potential errors or risks. The study explores strategies for fostering appropriate trust in generative AI tools, such as GitHub Copilot, to enhance their usability and reliability. & Ability, Benevolence, Integrity & \cite{Wang2024InvestigatingTrust} & \cite{Lee2004, Liao2022, Vereschak2021, EuropeanCommission2019, das2020opportunitieschallengesexplainableartificial, Boubin2017, OConnor2019, Pearce2022Asleep, Perry2023, Mayer1995} \\
\hline
6 & Trust involves ability (providing reliable performance), integrity (aligning with ethical and social norms), and benevolence (acting in society’s best interest). Trustworthiness is reinforced by verifying AI’s reliability, yielding ethical alignment, and strengthening oversight through public supervision and regulation, emphasizing governance over technical solutions. & Ability, integrity, benevolence, verifiability & \cite{Ganapati2024PublicValue} & \cite{DesouzaDawson2023, Mayer1995} \\
\hline
7 & Not formally, but conceptualizes it as the ability of AI systems to avoid generating false or misleading responses by accurately determining when they lack sufficient evidence to answer a given question. & Reliable, factual & \cite{ajewska2024} & - \\
\hline
8 & Trust in IT artifacts is defined by their quality, reliability, accuracy, and security. While organizational trust frameworks, such as Mayer et al. \cite{Mayer1995}, are widely used in IS research, applying concepts like benevolence to IT artifacts risks anthropomorphizing technology. & Ability/competence, integrity, and benevolence & \cite{Lacity2024} & \cite{Mayer1995, Porra2019, Belanger2019, Kim2009, Lowry2013} \\
\hline
9 & Not explicitly, but it describes trustworthy AI in terms of consistency, reliability, explainability, and safety. & Consistency, reliability, explainability, and safety & \cite{Gaur2024} & - \\
\hline
10 & ``Trust is an individual’s or group’s willingness to be vulnerable to another party based on the confidence that the latter party is benevolent, reliable, competent, honest, and open.'' This definition is applied to teacher-student relationships to explore how trust is built and eroded in an AI-mediated assessment environment. & Transparency, reliability, honesty, competence, integrity, fairness, benevolence, openness, trust-building & \cite{LuoJess2024} & \cite{Hoy1999} \\
\hline
11 & Believing in the credibility or accuracy of a statement. & Credibility, AI alignment & \cite{Buchanan2024} & - \\
\hline
12 & A willingness to be vulnerable, accompanied by a positive expectation of the other party’s intentions and behavior. Real-world AI incidents have urged the need for translating high-level trustworthiness rhetoric into domain-specific practices. This raises the question of how trust should be defined in fields such as the insurance industry and whether regulatory bodies should actively promote trust in AI systems. & AI ethics, AI trustworthy & \cite{Ressel2024} & \cite{Mayer1995} \\
\hline
13 & Not explicitly, but authors suggest that trust can be enhanced by providing indicators of familiarity, validation, or endorsements from experts. However, people tend to be more cautious when AI-generated content is explicitly labeled. & Transparency, disclosure, validation, familiarity, experience, task suitability, credibility & \cite{Califano2024} & - \\
\hline
14 & Trust is defined as possessing a confidence in something or someone. While personalization is key to fostering trust, the black-box problem -- lack of transparency and explainability -- hinders trust. Trustworthy AI should not only be lawful, ethical, and robust, but also give humans the ability to have control over its use, draw from and provide reliable data, offer security and privacy to its users, and be transparent, accountable, and inclusive. Trust in AI-LLM means that individuals and institutions are confident that the data used by, and information generated in, AI-LLM are impartial. & Confidence, fair, unbiased, ethical, lawful, robust, human oversight, transparency, accountability, security, privacy & \cite{Jenks2024} & \cite{cook2001trust, Araujo2020, EuropeanCommission2025} \\
\hline
15 & Trust is a multidimensional construct that encompasses several key components, including benevolence, competence, and integrity. Users tend to trust AI-generated content less when they are aware of its origin. & Credibility, Transparency & \cite{JasperJia2025} & \cite{Mayer1995} \\
\hline
16 & Trust is defined as the belief that an agent will assist in achieving an individual’s goals in situations marked by uncertainty and vulnerability. Addressing trust is fundamental from early deployment stages, as initial errors can have long-term consequences on user trust calibration and later adoption.  & Transparency, calibrated trust, reliability & \cite{Martell2024} & \cite{Mayer1995, Lee2004, Yang2020, Aroyo2021, Hoff2014, Dietz2006, Toreini2020} \\
\hline
17 & Trust in generative AI can be divided into cognitive trust and affective trust. Cognitive trust is based on a user’s evaluation of the AI platform’s ability to deliver accurate and reliable information, while affective trust stems from the user’s belief that the AI system considers their interests and well-being. & Reliable, credible & \cite{Zhou2024} & \cite{Zhang2014, Wang2016, Stewart2003} \\
\hline
18 & The author actually criticizes the term trustworthy and ethical AI because they can humanize AI. & Responsible AI & \cite{BaezaYates2024} & - \\
\hline
\end{tabular}%
}

\end{table*}

In Table \ref{tab:trust_definitions}, we provide a \textbf{structured overview of various definitions and conceptualizations of trust and trustworthiness in AI}, as discussed in the literature. This compilation allows for a comparative analysis of how trust in AI is framed across different domains, emphasizing dimensions such as ability, reliability, integrity, explainability and ethical alignment. From the initial 68 analysed papers, 18 papers are explicitly or implicitly define trust/trustworthiness in LLM or genAI. That is, 26.47\% against 73.53\% of papers that are not providing useful information for this task. 50 papers were not taken into consideration because they are either closed access, on the use of LLM to improve trust in a different domain e.g., radiology, measuring trust level of a certain group when using certain LLM system, or simply with a different focus.

%The complete analysis is available on Zenodo.

We found a tendency in reviewed studies in utilizing Mayer et al.'s work --  ``An Integrative Model of Organizational Trust'', published in 1995 -- to build on definitions for what is trustworthiness in LLM -- 33,33\%, one third, of definitions (6/18 entries: 5 \cite{Wang2024InvestigatingTrust}, 6 \cite{Ganapati2024PublicValue}, 8 \cite{Lacity2024}, 12 \cite{Ressel2024}, 15 \cite{JasperJia2025}, 16 \cite{Martell2024}) explicitly draw from Mayer et al.’s \cite{Mayer1995} organizational trust framework. Trust is described as a three factor construct consisting of ability, benevolence, and integrity \cite{Mayer1995}. 
Martell et al. \cite{Martell2024} add predictability on Mayer et al.'s work, naming this as ABI+ factors of perceived trustworthiness towards AI:

\begin{enumerate}
    \item \textbf{Ability:} AI's capacity to perform tasks effectively,
    \item \textbf{Benevolence:} AI’s intent to act in users' best interests,
    \item \textbf{Integrity:} adherence to ethical standards,
    \item \textbf{Predictability:} consistency in AI responses.
\end{enumerate}

Trust in AI, particularly in situations of uncertainty and vulnerability, is described as the belief that an AI agent will support an individual in achieving his or her goals \cite{Wang2024InvestigatingTrust}. Trust is fundamentally linked to vulnerability and uncertainty, as highlighted by multiple perspectives. 

Hoy and Tschannen-Moran \cite{Hoy1999} define trust as a willingness to be vulnerable based on confidence in another party’s benevolence, reliability, competence, honesty, and openness. Mayer et al. \cite{Mayer1995} reinforce this by framing trust as a positive expectation of another’s intentions and behavior, emphasizing the risk involved in relying on an entity. Similarly, Lee and See \cite{Lee2004} describe trust as the belief that an agent will assist in achieving one’s goals despite uncertainty and vulnerability, underscoring that trust in AI systems depends not just on their performance but also on users' confidence in their intent and reliability. In this context, users assume the vulnerable position of relying on genAI models or proprietary platforms with the expectation that this risk will be beneficial to them, i.e. that it will help them achieve their goals despite possible misinformation or bias \cite{Ressel2024}. 

In the paper ``Mitigative Strategies for Recovering From Large Language Model Trust Violations'' \cite{Martell2024}, the authors present three different levels of trust calibration, as shown in Table \ref{tab:trust-calibration}.

\begin{table}[h]
    \centering
    \renewcommand{\arraystretch}{1.1}
    \caption{Types of Trust Calibration in Human-AI Interaction \cite{Martell2024}}
    \label{tab:trust-calibration}
    \resizebox{\columnwidth}{!}{%
    \begin{tabular}{p{9.5cm}l}
        \toprule
        \textbf{Concept and Definition} & \textbf{Reference} \\ 
        \midrule
        \textbf{Appropriate (Calibrated) Trust}: Alignment between an individual's perceived trust level and the AI system's actual capabilities and performance. & \cite{Yang2020} \\
        
        \textbf{Over-Trust}: Occurs when users:
        \begin{itemize}
            \item Perceive the AI as more capable than it truly is
            \item Treat the technology as a fully competent teammate
            \item Over-rely on AI outputs despite known limitations
        \end{itemize} & \cite{Aroyo2021} \\

        \textbf{Under-Trust}: Occurs when users:
        \begin{itemize}
            \item Underestimate the AI system's capabilities
            \item Over-rely on their own judgment
            \item Dismiss potentially valuable AI contributions
        \end{itemize} & \cite{Hoff2014} \\
        \bottomrule
    \end{tabular}%
    }
\end{table}

The same study highlights the need to prioritise accuracy in LLM deployment, as early errors can irrevocably damage user confidence, calibration and subsequent adoption. In other words, operationalising trustworthiness in AI-LLM systems from the early stages is fundamental, as it is difficult to regain user trust. This can be seen in cases such as Microsoft's Tay Bot \footnote{\href{https://blogs.microsoft.com/blog/2016/03/25/learning-tays-introduction/}{https://blogs.microsoft.com/blog/2016/03/25/learning-tays-introduction/}}. This AI-powered chatbot started tweeting offensive neo-Nazi comments within hours of its release and was subsequently shut down.

In Table \ref{tab:trustworthiness_terms}, we present the frequency of key terms associated with trustworthiness in AI, reflecting the most commonly emphasized aspects in the literature. This Table was produced by analysing the synonyms to trustworthy AI presented in Table \ref{tab:trust_definitions}. 

    \begin{table}[h]
    \centering
    \caption{Frequency of Trustworthiness-Related Terms Extracted from Reviewed Papers (2019-2025)}
    \label{tab:trustworthiness_terms}
    \renewcommand{\arraystretch}{1.3}
    \resizebox{3cm}{!}{%
    \begin{tabular}{lc}
        \toprule
        \textbf{Term} & \textbf{Frequency} \\
        \midrule
        Transparency    & 12 \\
        Explainability  & 9  \\
        Reliability     & 9  \\
        Ethics          & 6  \\
        Privacy         & 6  \\
        Fairness        & 5  \\
        Responsible     & 5  \\
        Credibility     & 4  \\
        Robustness      & 4  \\
        Trustworthy     & 4  \\
        \bottomrule
        
    \end{tabular}%
    }
\end{table}

%%%%%%%%%%%%%%%%%%%%%%%%%%%%%%%%%%%%%%%%%%%%%%%%%%%%%%%%%%%%%%%
%\subsection{Practical LLM trustworthiness-enhancing techniques found in the literature}

% Here was RQ3

%Most techniques (12/20) target the post-training phase (e.g., fine-tuning, RAG, XAI methods), reflecting a reactive focus on mitigating emergent risks rather than preempting them during pre-training.
From the 68 documents analysed, 19 unique documents contribute to the creation of the Table \ref{tab:trust_enhancing}, which provides \textbf{strategies for enhancing trustworthiness in LLM.} We found 20 different strategies, categorised by their LLM lifecycle phase -- either pre-training, post-training or inference phase -- the actor responsible for their implementation -- either developer or user -- and the trustworthiness attributes associated with them -- e.g. transparency, explainability, reliability.

\begin{table*}[!t]
\centering
\caption{Trustworthiness-Enhancing Strategies in LLMs}
\label{tab:trust_enhancing}
\renewcommand{\arraystretch}{1.2}
\resizebox{\textwidth}{!}{%
\begin{tabular}{|c|p{3.2cm}|p{2.5cm}|p{2.5cm}|p{4.2cm}|p{2.7cm}|}
\hline
\textbf{\#} & \textbf{Strategy} & \textbf{LLM Lifecycle Phase} & \textbf{Responsibility} & \textbf{Trustworthiness Attributes} & \textbf{Citations} \\
\hline
1  & (Supervised) Fine-tuning & Post-training & Developer & Reliability, Robustness, Fairness & \cite{Jha2023Challenges, Kilhoffer2024, Barman2024, Ressel2024, Chen2024, Roumeliotis2025} \\
\hline
2  & RAG & Post-training & Developer & Faithfulness, Transparency, Reliability & \cite{Jha2023Challenges, Lan2024, Ressel2024, Hannah2025} \\
\hline
3  & Training & Pre-training & Developer & Robustness, Accuracy & \cite{Bhattacharya2023} \\
\hline
4  & Human-in-the-loop & Inference & User & Fairness, Transparency, User Trust & \cite{Wang2024, Chen2024} \\
\hline
5  & Stakeholder feedback & Inference & User & Trust Calibration, Social Acceptability & \cite{Wang2024} \\
\hline
6  & Prompt engineering & Inference & User & Safety, Controllability, Fairness & \cite{Mousavi2024, Hannah2025} \\
\hline
7  & Evaluation & Post-training & Developer & Accountability, Transparency & \cite{Mousavi2024, Nag2023} \\
\hline
8  & Contextual Explanation & Pre-/Post-training & Developer & Explainability, Transparency & \cite{Chari2023} \\
\hline
9  & In-context Learning (ICL) & Post-training & Developer & Adaptability, Efficiency, Fairness & \cite{Zhang2024, Barman2024} \\
\hline
10 & RAG-Ex Framework & Post-training & Developer & Explainability, Transparency & \cite{Sudhi2024} \\
\hline
11 & Uncertainty Expression & Post-training & Developer & Transparency, Explainability, Reliability & \cite{Kim2024} \\
\hline
12 & Self-explanations & Post-training & Developer & Explainability, Interpretability & \cite{Kim2024} \\
\hline
13 & Post-release audits & Post-training & Developer & Transparency, Accountability, Explainability, Ethics & \cite{FernandezNieto2024} \\
\hline
14 & RLHF & Inference & Developer/User & Alignment & \cite{Barman2024, Chen2024} \\
\hline
15 & Chain of Thought (CoT) & Inference & Developer & Explainability & \cite{Barman2024, Chen2024} \\
\hline
16 & Blockchain & Post-training & Developer & Security, Transparency, Tamper-resistance & \cite{Fan2024} \\
\hline
17 & Knowledge Graph & Post-training & Developer & Transparency, Explainability & \cite{Chen2024, Hannah2025} \\
\hline
18 & Data curation & Pre-training & Developer & Fairness, Bias Mitigation & \cite{Chen2024} \\
\hline
19 & External Tools (e.g., search engines) & Inference & Developer & Explainability & \cite{Chen2024} \\
\hline
20 & XAI techniques (LIME, SHAP, etc.) & Post-training & Developer & Explainability, Interpretability, Transparency & \cite{Mersha2025} \\
\hline
\end{tabular}%
}
\end{table*}

In the first place, with 6 different documents referring to this strategy, is (supervised) fine-tuning (SFT), in the post-training phase. This strategy is the responsibility of the LLM system developer and focuses mainly on reliability, robustness and fairness. SFT has enabled models to become helpful assistants, that is, only the pre-trained models themselves were not the desired chatbot that ChatGPT came to be. 
%Nowadays, there is extensive documentation such as \href{https://platform.openai.com/docs/guides/fine-tuning}{OpenAI fine-tuning docs}, and platforms such as \href{https://colab.research.google.com/}{Google Colab}, which have facilitated the availability of hardware and environment to allow fine-tuning by anyone anywhere in the world.

In second place, with 4 different documents referring to this strategy, is Retrieval Augmented Generation (RAG), which also takes place in the post-training phase and under the responsibility of the developer. This technique has proved to be efficient, cost-effective and allows models to base their responses on frequently updated documents. This strategy has demonstrated versatility and requires fewer technical resources to implement. The trustworthiness attributes most associated with RAG are faithfulness, transparency and reliability.

Developers have primary responsibility for trustworthiness (16/20 techniques), while users contribute minimally (4/20, e.g., human-in-the-loop feedback). This raises concerns about responsibility balance, as the liability of positive trustworthiness outcomes relies mostly in the developers hands. Users can improve the trustworthiness of AI-LLM systems mainly in the inference phase, through human-in-the-loop (evaluating the output), stakeholder feedback (expert feedback), prompt engineering (implementing prompt techniques) and Reinforcement Learning from Human Feedback (RLHF) (providing feedback/labelling to the generated output).

%Only 3 methods address pre-training (data curation, training, contextual explanation), suggesting underdeveloped preventive approaches.
Most techniques (12/20) target the post-training phase (e.g. fine-tuning, RAG, XAI methods). Regarding trustworthiness attributes, explainability and transparency come out on top (14/20 entries), driven by techniques such as uncertainty expressions (Entry 11) and knowledge graph (Entry 17). In contrast, security (Entry 16: blockchain) and ethics (Entry 13: audits) receive limited attention.

%%%%%%%%%%%%%%%%%%%%%%%%%%%%

\section{Limitations}
\label{limitations}

While bibliometric analysis provides valuable insights into research trends, it has notable limitations. Its emphasis on publication and citation counts does not always capture the depth, quality or maturity of a field of research \cite{Donthu2021HowTCbibliometric}. High citation rates can indicate popularity rather than true intellectual impact, and biases in citation patterns can distort impact \cite{Donthu2021HowTCbibliometric}.

To address these concerns, this study combines bibliometric techniques with qualitative assessments of articles with high focus on discussing trustworthiness (i.e., studies with ``trust*'' in the abstract). By integrating both perspectives, we aim to provide a more comprehensive understanding of LLM trustworthiness research, including also more detailed, qualitative analysis of select papers in addition to the quantitative bibliometric approach.

Another limitation of this study is that it relies solely on data from the Web of Science database. This results in two limitations. First, given the rapid pace of research in LLMs, a significant proportion of relevant studies are often published on arXiv, a widely used open-access preprint repository maintained by Cornell University. By not including arXiv, this study may not capture the latest developments and trends in the field. However, by not including them, this study aims to focus only on peer-review papers. Second, by only focusing on one database, we will also have excluded some potentially relevant, peer-reviewed and already published research. However, this was done due to technical limitations related with the chosen bibliometric approach.

%%%%%%%%%%%%%%%%%%%%%%%%%%%

\section{Discussion and Conclusions}
\label{discussion_and_conclusions}

\textit{RQ1: What are the research trends, key authors, and thematic areas in the study of LLM trustworthiness?}

% Anual Publication Per Year

As shown in Figure \ref{fig:annualscientific_production}, the sharp increase in publications after 2022 can be attributed to advances such as GPT-3/4, regulatory debates (e.g., EU AI Act), and societal concerns about LLM trustworthiness and ethical outcomes. This highlights the increase interest researchers have in trust/trustworthiness and ethical LLM and genAI research. This perceived trend is expected to continue, reflecting the growing attractiveness of the field to both established researchers and newcomers.

% Most Global Cited Documents

Figure \ref{fig:most_Global_cited_documents} shows articles with high global citations, indicating a significant impact and influence across disciplines \cite{Donthu2021HowTCbibliometric}. Kasneci et al. \cite{Kasneci2023} explore the opportunities and challenges of using LLMs like ChatGPT in education, highlighting their potential to personalize learning, create educational content, and support both students and teachers. However, it also addresses critical risks such as bias, over-reliance, ethical concerns, and sustainability, emphasizing the need for responsible integration, teacher training, and ongoing research. The authors advocate for a balanced approach that uses LLMs to improve education while providing ethical use, transparency, and equitable access. Ongoing efforts to perceive the use of LLMs in \textbf{education} highlights how academia is concerned about this disruptive tool in educational settings.

Dwivedi et al. \cite{Dwivedi2023} explores the \textbf{multidisciplinary impact of ChatGPT}, highlighting its potential to enhance productivity across industries while raising concerns about ethics, misinformation, and transparency. It discusses its role in \textbf{research, business, and education}, emphasizing both opportunities and challenges, including AI's effect on academic integrity and hybrid work. The authors propose multiple future research directions, e.g., on AI governance, human-AI collaboration, and responsible implementation.

Sallam Malik \cite{Sallam2023} systematically reviews the potential applications and limitations of ChatGPT in \textbf{healthcare}. It highlights the benefits of ChatGPT in \textbf{scientific writing, research efficiency, medical education, and clinical workflows}, while addressing concerns such as misinformation, bias, ethical and legal challenges, and safety risks. The study emphasises the need for responsible use of AI in healthcare and calls for regulatory guidelines and stakeholder collaboration to ensure ethical and effective implementation.

The three highly cited documents analysed present investigations of the specific adoption and impact of ChatGPT in different domains -- \textbf{education} \cite{Kasneci2023}, \textbf{research, business, and education} \cite{Dwivedi2023}, and \textbf{healthcare} \cite{Sallam2023}. This shows how the research community has responded to the rapid emergence of ChatGPT in several domains. Furthermore, all three top documents have in common the need to ensure the ethical use, transparency and responsible implementation of AI.

%%%
% Thematic Evolution

The results shown in Figure \ref{fig:thematic_evolution} indicate a progressive shift in the discourse from general AI ethics toward more specialized concerns about LLM trustworthiness. The focus on ‘AI ethics’ (2023-2024) has shifted to concerns about ‘disinformation’ (2025-2025), reflecting the fact that LLMs are rapidly replacing traditional search engines and are primarily used as chatbots, particularly in question-answering (QA) settings. As users increasingly rely on LLMs rather than traditional search engines to retrieve information, the risk of misinformation and manipulated content has become a central issue. This typical type of use of LLMs suggests why discussions of AI ethics have evolved to disinformation.

The term ‘trustworthiness’ (2025-2025) stems from the term ‘responsible AI’ (2023-2024), suggesting the discourse shift from traditional responsible AI principles to trustworthiness in genAI. This suggests that responsible AI discussions have been rebranded and adapted to meet the LLM challenges. Moreover, AI ethics research focus has shifted from abstract guiding principles -- e.g., \cite{jobin2019globallandscape, Correa2023WorldwideAIethics200, hagendorff2020ethics, Ryan2020ArtificialIE}. -- to abstract LLM trustworthiness taxonomies -- e.g., \cite{sun2024trustllm, Wang2023DecodingTrustAC, liu2023trustworthy}.

Researchers unfamiliar with ethical aspects in AI are being attracted to explore it but in the LLM and genAI hype. Thus, they are coming up with terminologies and theoretical approaches, that can be seen as yet another set of AI ethics guidelines. This abrupt interest is possibly hindering the actual advancements towards more ethical AI, as it is masking the debate towards a ``new'' set of theoretical guidelines, but not providing practical guidance for the developers, the ones ultimately responsible for the development of ethical AI or trustworthy LLM \cite{fromwhattohow2019floridi}. In other words, this perceived rebranding of responsible AI and AI ethics to trustworthiness and disinformation is for the most part failing in proposing useful unified terminologies, frameworks, or practical guidance for developers to operationalise principles.

There is a growing need for enforceable guidelines beyond non binding ethical commitments. Even more evident after big tech companies such as Google withdrawing its AI ethical principles in mass surveillance and AI weapons \cite{google2025endingaiban}. Consequently, there is an urge for actual hard laws that can govern the development and use of these systems, that can move beyond the EU AI Act \cite{sillberg2024euaiactgood}, due to the fact that it only propose risk level assessments, and links to The European Commission’s High-level expert group on AI (AI HLEG) \cite{TrustworthyAIHLEG}. The later has a set of trustworthy AI principles, however, it is non binding. Despite the fact that a binding document links to a non binding one, does not automatically transform the AI HLEG into hard law. On top of that, an unified set of practical guidance are still lacking, after several years of debate -- at least since 2019 \cite{fromwhattohow2019floridi}.

Despite this progress, fragmentation remains, with multiple authors competing with definitions of LLM trustworthiness rather than a single, unified framework. The continued proliferation of non-binding framework proposals and terminologies by singular researchers highlights the urgency of bridging the gap between theoretical AI ethics and its real-world implementation through unified frameworks and legislation. This observation goes in line with tech ethics limitations identified by Green's \cite{Green2021Contestation} where a vagueness and focus on individuals leads to ethics washing, stakeholders ease public voice and resist to governmental legislation by embracing the ethical discourse, without real commitment.

%%%
% Word Cloud

Figure \ref{fig:word_cloud} shows the most frequent keywords that appeared in the analysed literature on LLMs, genAI, trust and ethics. ‘ChatGPT’ is the most recurrent term as it dominates as the most popular LLM in the literature, while ‘health’ and ‘education’ are the domains most explored. Furthermore, ‘privacy’, ‘ethics’, ‘bias’ are the most relevant regarding ethical issues. This indicates that most of the studies are investigating how privacy and bias are the most explored terms in health and education through the use of ChatGPT.

%%%
% Coupling Map

From the coupling map, the studies identified in Figure \ref{fig:clustering} and Table \ref{tab:top_authors} highlight the applications of LLMs across multiple domains and the ethical, practical, and evaluative challenges. Martinez-Maldonado and colleagues \cite{Yan2023} systematically review the use of LLMs in education and identify nine key applications, such as grading, content generation and feedback, while cautioning about replicability, transparency and ethical concerns. Park and co-authors \cite{Kim2023} investigate the impact of misinformation generated by ChatGPT on tourism decision-making, demonstrating how trust in AI can be easily manipulated when incorrect information is presented. Similarly, Sallam \cite{Sallam2023} examines AI in healthcare, pinpointing its potential to improve medical research, education and clinical workflows, as well as its potential risks such as bias, hallucination, plagiarism and cybersecurity threats.

Wang et al. \cite{chang2023survey} provide a meta-analysis of AI evaluation methods and argue for a more structured approach to assessing their performance, ethical considerations and real-world impact in different domains. Herrera et al. \cite{Longo2024} explore XAI, pointing the urgent need for transparency in AI decision-making and proposes 28 open problems for achieving better interpretability and accountability in AI systems. 

Overall, these authors are exploring trust in domains involving education, tourism, healthcare, stressing how LLMs are reshaping a multitude of sectors in everyday life. Furthermore, the studies emphasize how evaluation methods and transparency are closely related to trust.

%%%

\textit{RQ2: What are the LLM trust/trustworthiness definitions proposed by different authors in the literature?}

In an attempt to find LLM trustworthiness definitions from the literature, we devised Table \ref{tab:trust_definitions}, from which we point that several authors are building on Mayer et al.'s \cite{Mayer1995} organizational trust model, indicating a deficiency in the development of a new framework specifically made for AI and LLM systems. While Mayer's model is broadly adopted in academia, it was devised before the widespread adoption of AI technologies (1995). Thus, adapting a 30-year-old framework to the AI era raises important questions. Similarly to AI ethics, where soft law recommendations did not pose serious changes, it is worth discussing whether trustworthiness in LLMs should instead be governed by hard laws that unify terms and recommendations.

The concept of trust calibration -- especially the risk of over-trust and under-trust described by Martell et al. \cite{Martell2024} -- reinforces the importance of addressing trustworthiness early in the deployment process. As the case of Microsoft's Tay Bot demonstrates, early missteps can permanently damage user trust, making recovery almost impossible. The findings also highlight that users are inherently vulnerable when interacting with LLMs and generative AI systems, as they rely on these systems with the expectation of goal achievement despite possible misinformation or bias.

Furthermore, the more familiar we are with a tool, the more we tend to trust them \cite{Califano2024}. Other authors state that trust in AI-LLM can be translated as the degree to which users and institutions are confident that the data used by, and the information generated by, AI-LLM is impartial, i.e. unbiased \cite{Jenks2024}.

Two more trust related dimensions are presented in Table \ref{tab:trust_definitions}: cognitive trust (rational evaluation of a system’s reliability and accuracy) and affective trust (emotional confidence in its benevolence and alignment with user interests) \cite{Zhang2014}. Cognitive trust, anchored in perceived credibility, facilitates affective trust through sustained reliance, as users transfer trust from the platform's technical competence to its perceived ethical intent \cite{Zhang2014, Wang2016}. This interdependence highlights how platform reliability shapes users' perceptions of AI-generated content, with technical accuracy (‘reliable’) and empathetic design (‘credible’) jointly calibrating holistic trust.

%What are the trade-offs of using or adapting a 30 year old organizational trust framework to the AI era? First, as a benefit, it is a well established set of trust taxonomy, that has been widely accepted in the academia. However, it was devised way before wider adoption of different AI technologies, in special LLMs. Similarly to AI ethics, where soft law recommendations did not pose serious changes, should trustworthiness in LLM discussion be on hard laws that can unify the terms and recommendations?

In \cite{Jenks2024}, the authors state three principles of AI-LLM: fair, unbiased, and ethical. On top of that, they relate these principles with the work done within the European Commission on AI ``Building Trust in Human-Centric Artificial Intelligence'' \cite{EuropeanCommission2019}, which states that trustworthy AI should not only be lawful, ethical, and robust, but also give humans the ability to have control over its use, draw from and provide reliable data, offer security and privacy to its users, and be transparent, accountable, and inclusive. The efforts of the European Commision on AI is better translated in the EU AI Act \cite{sillberg2024euaiactgood, europaActFirst_EUAIAct}, where a risk level assessment for AI systems is provided. We briefly describe the EU AI Act's risk level in Table \ref{tab:euaiact_risk_levels}. 
%From this Table, we pinpoint that most of genAI and LLM generated content falls under the ``Limited Risk'' level, where there is a need for Transparency requirements, highlighting once again the importance of this trustworthy principle.

\begin{table}[h]
    \centering
    \caption{Risk Levels in the EU AI Act}
    \label{tab:euaiact_risk_levels}
    \renewcommand{\arraystretch}{1.3}
    \resizebox{\columnwidth}{!}{%
    \begin{tabular}{ll}
        \toprule
        \textbf{Risk Level} & \textbf{Description} \\
        \midrule
        Unacceptable Risk & AI systems that are prohibited due to threats to fundamental rights, e.g., social scoring, manipulation. \\
        \addlinespace
        High Risk & AI systems that impact safety or fundamental rights, e.g., biometric ID, credit scoring, law enforcement. \\
        \addlinespace
        Limited Risk & AI systems with transparency requirements, e.g., chatbots, AI-generated content. \\
        \addlinespace
        Minimal Risk & AI systems with no specific regulations, e.g., spam filters, AI-driven recommendations. \\
        \bottomrule
    \end{tabular}%
    }

\end{table}

In addition, within the EU AI Act there is reference to a non-binding document, also provided by the European Commission, that aims to provide a set of trustworthy principles in AI. This non-binding document is the 2019 Ethics Guidelines for Trustworthy AI developed by the AI High-Level Expert Group (AI HLEG) \cite{TrustworthyAIHLEG}. We shortly present its principles in Table \ref{tab:hleg_principles}. This Table reinforces terms already discussed in Table \ref{tab:trustworthiness_terms}, showing a level of alignment between academic and policy perspectives.
    
    \begin{table}[h]
    \centering
    \caption{AI HLEG Principles for Trustworthy AI -- 2019}
    \label{tab:hleg_principles}
    \renewcommand{\arraystretch}{1.3}
    \resizebox{\columnwidth}{!}{%
    \begin{tabular}{ll}
        \toprule
        \textbf{Principle} & \textbf{Description} \\
        \midrule
        Human Agency and Oversight & AI should respect human autonomy and be controllable. \\
        Technical Robustness and Safety & AI should be resilient, secure, and function as intended. \\
        Privacy and Data Governance & AI must ensure data protection and quality. \\
        Transparency & AI should be explainable and provide clear information. \\
        Diversity, Non-discrimination, and Fairness & AI should be inclusive and avoid bias. \\
        Societal and Environmental Well-being & AI should promote sustainability and benefit society. \\
        Accountability & AI should have mechanisms for responsibility and oversight. \\
        \bottomrule
    \end{tabular}%
    }
\end{table}

Finally, in \cite{BaezaYates2024}, there is a critic towards the use of the terms trustworthy and ethical AI, as they can be used to humanize, i.e., anthropomorphise AI. This critic is valid insofar every Information System artefact, such as LLM and AI systems, are devised by humans, and they are programmatically designed to fulfil a set of subjective or objective desires implied by the developers or organizations they are inserted.

In Table \ref{tab:trustworthiness_terms} Transparency and Explainability appear as the first and second most frequently mentioned characteristics, respectively, highlighting how they are enabling principles, i.e., without transparency and explainability it is impossible for other trustworthiness or ethical principles to exist \cite{siqueira2021ethical}. Reliability is related to Ability and Predicability in the ABI+ framework \cite{Mayer1995}. Furthermore, Ethics, Privacy, and Fairness highlight concerns related to Integrity, i.e., adherence to ethical standards.

%%%

\textit{RQ3: What practical techniques and tools can be used to enhance LLM trustworthiness throughout the LLM lifecycle?}

The techniques most frequently mentioned in the literature, as shown in Table \ref{tab:trust_enhancing} are SFT and RAG, which stand out for their technical feasibility. Furthermore, most of the strategies identified are post-training. This overemphasis on post-training strategies, rather than addressing issues earlier in the pre-training phase (e.g., bias mitigation through pre-training data curation), should not necessarily be interpreted as a lack of concern for preventive risk mitigation. On the contrary, it reflects the resource-intensive nature of the pre-training phase, including data collection and model training. In addition, the developer is responsible for most of the strategies, highlighting the crucial role of practitioners in operationalising trustworthy LLM and the need to provide them with practical guidance.

Regarding the distribution of trustworthiness attributes identified, it indicates a misalignment with regulatory priorities such as the risk levels of the EU AI Act, e.g., security appearing only once. Further, the strategies identified suggest the need for integrated frameworks that unify technical interventions (e.g. RAG, knowledge graphs) with governance mechanisms (e.g. post-release audits, human-in-the-loop) across the LLM lifecycle, with a strong adherence to legislations. From a policy perspective, efforts such as the EU AI Act are steps in the right direction, but there remains an urgent need for stronger regulations that mandate transparency and accountability in the development and deployment of LLMs.

%%%%%%%%%%%%%
% Conclusion

In conclusion, despite the growing body of literature on LLM trustworthiness, there is still no unified set of practical guidelines to help developers of LLM-based systems mitigate risks throughout the development and deployment process. To bridge this gap, we conducted a bibliometric analysis and manual reading of papers related to trustworthiness in LLMs, highlighting research trends, definitions and practical guidance. The findings reveal a shift from traditional AI ethics discussions towards trustworthiness frameworks, but a lack of consensus and practical implementation strategies persists. Most common terms related to trustworthiness extracted from reviewed papers are transparency, explainability, reliability, ethics and privacy. While various technical strategies were idenified, enforceable regulations and standardised frameworks remain necessary for real-world impact. Multiple authors are competing for LLM trustworthiness terminologies rather than unified frameworks and legislation, this fragmentation and vagueness leads to risks of ``ethics washing,'' where stakeholders ease public voice and resist to legislation by embracing the ethical discourse, without real commitment. 

Future work includes implementing and evaluating individual trustworthiness-enhancing strategies in LLMs identified. Moreover, different combinations of technical strategies, such as RAG, knowledge graph, XAI and fine-tuning, would shed some light on the trustworthiness practices. Furthermore, adequate training and practical guidance for developers in the operationalisation of trustworthiness attributes should be provided. In addition, interdisciplinary collaboration is essential to reconcile ethical principles with technical and regulatory considerations, thus involving potential stakeholders as well as users of the system. Ultimately, progress towards trustworthy AI will require not only technological implementations, but also unified legislation, practical frameworks, public and industry engagement to ethical AI development.

\section*{Acknowledgment}

This research was supported by Jane and Aatos Erkko Foundation through CONVERGENCE of Humans and Machines Project under grant No. 220025. We acknowledge Muhammad Waseem for his thoughtful contribution to the revision process through detailed and constructive feedback.

\bibliographystyle{ieeetr}
\bibliography{reference}

\end{document}